\documentclass[10pt,twocolumn,letterpaper]{article}

\usepackage{cvpr}
\usepackage{times}
\usepackage{svg}
\usepackage{epsfig}
\usepackage{graphicx}
\usepackage{amsmath}
\usepackage{amssymb}
\usepackage{float}
\usepackage[pagebackref,breaklinks,colorlinks]{hyperref}
\newif \ifproofread

\begin{document}

\title{AdvGen: Physical Adversarial Attack on Face Presentation Attack Detection Systems}

\author{$\text{Sai Amrit Patnaik}^1,\ \text{Shivali Chansoriya}^1,\ \text{Anoop M. Namboodiri}^1\ \text{and}\ \ \text{Anil K. Jain}^2$\\
$^1\text{IIIT Hyderabad, India},\ \ ^2\text{Michigan State University, USA}$\\
{{\tt \small \{sai.patnaik, shivali.chansoriya\}@research.iiit.ac.in, \ \tt\small anoop@iiit.ac.in}, \ {\tt\small jain@cse.msu.edu} \ }
}


\newcommand{\modelname}{AdvGen }
\newcommand{\sota}{state-of-the-art }
\proofreadtrue

\maketitle
\thispagestyle{empty}


\begin{abstract}


   Evaluating the risk level of adversarial images is essential for safely deploying face authentication models in the real world. Popular approaches for physical-world attacks, such as print or replay attacks, suffer from some limitations, like including physical and geometrical artifacts. Recently adversarial attacks have gained attraction, which try to digitally deceive the learning strategy of a recognition system using slight modifications to the captured image. While most previous research assumes that the adversarial image could be digitally fed into the authentication systems, this is not always the case for systems deployed in the real world. This paper demonstrates the vulnerability of face authentication systems to adversarial images in physical world scenarios. We propose AdvGen, an automated Generative Adversarial Network, to simulate print and replay attacks and generate adversarial images that can fool \sota PADs in a physical domain attack setting. Using this attack strategy, the attack success rate goes up to 82.01\%. We test AdvGen extensively on four datasets and ten state-of-the-art PADs. We also demonstrate the effectiveness of our attack by conducting experiments in a realistic, physical environment.

\end{abstract}


\section{Introduction}\label{introduction}
\begin{figure}[t]
  \centering
   \includegraphics[width=1.0\linewidth]{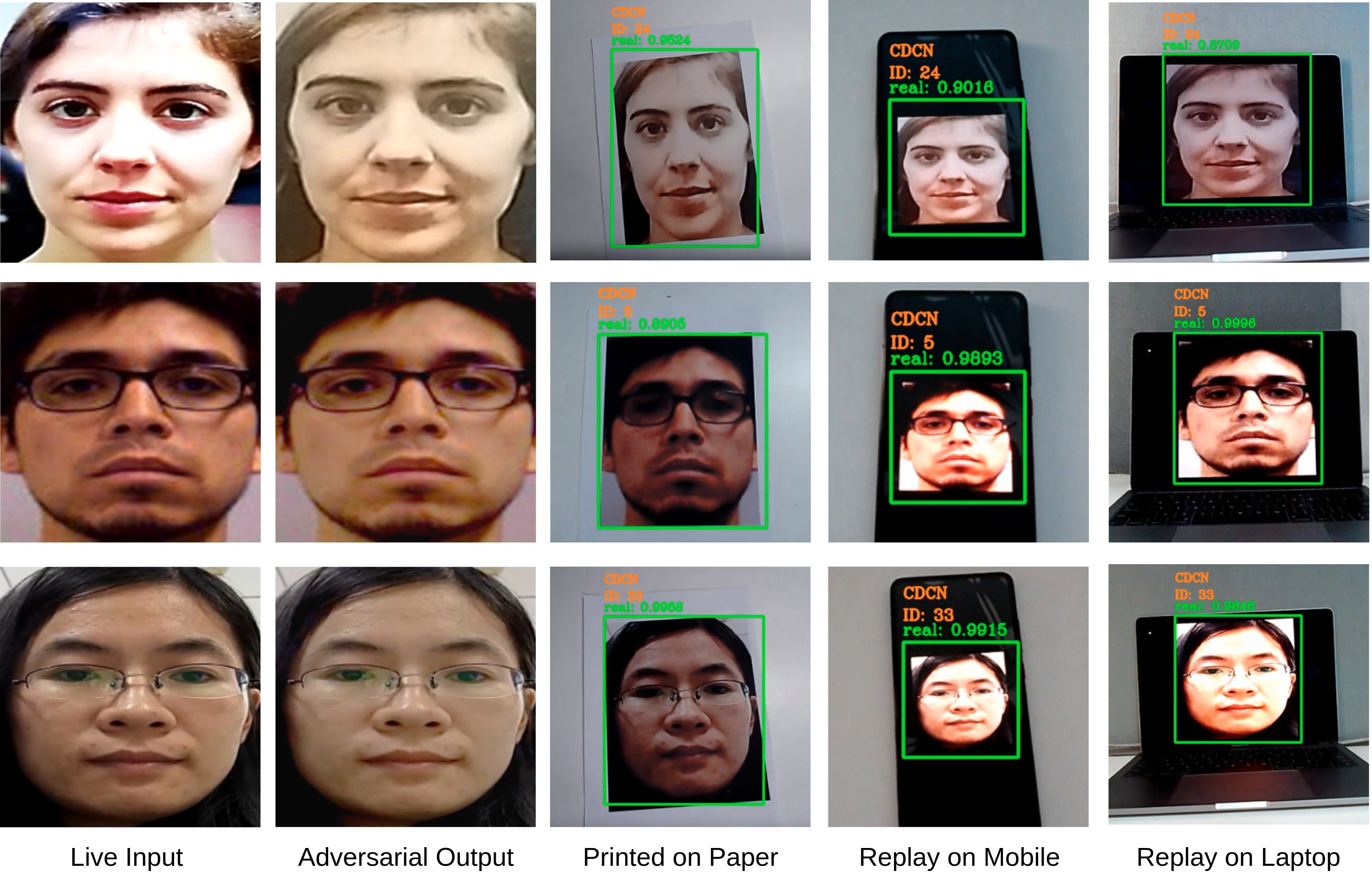}
   \caption{Example live images and corresponding adversarial images generated by AdvGen. First Column: live images from presentation attack datasets, second column: the corresponding adversarial images generated by AdvGen, third column: the predicted class along with the confidence score and recognized identity for a generated image(presenting an adversarial image generated by our model to the face recognition, fourth column: replay attack on a mobile screen, fifth column: replay attack on a laptop screen. The proposed method generates visually indistinguishable adversarial images from the input that is robust to distortions introduced after physical transformations.}
   \label{fig:finalresult}
\end{figure}
Face recognition systems are extensively used in real-time applications, such as surveillance systems, forensics, automated border control, user authentication~\cite{8119148}, payment processing, and security control systems. To prevent unauthorized access and attacks, Presentation Attack Detectors (PADs) are integrated into these systems (Figure \ref{fig:fig4}) to detect and reject presentation attacks, such as print attacks and replay attacks. As presentation attacks try to bypass the authentication system, understanding and correcting the potential pitfalls of a PAD module is as essential as designing high-accuracy recognition algorithms.

Most of the current \sota approaches use auxiliary information \cite{yu2020searching, atoum2017face, yu2022deep} to improve the performance and generalizability of the presentation attack detectors. Presentation and adversarial attacks on face recognition systems are still a significant concern. In a presentation attack, attacks are created using printed photographs, replayed videos, wearing a mask or makeup, etc. For generating presentation attacks, the hacker must actively participate by wearing a mask or replaying a photograph/video of the genuine individual, which may be conspicuous in scenarios involving human operators. Adversarial attacks, on the other hand, do not require active participation during verification.

The use of deep learning has significantly improved the accuracy of Presentation Attack Detectors. Adversarial attacks~\cite{szegedy2013intriguing, goodfellow2014explaining, moosavi2017universal, dong2018boosting}, however, exploit the vulnerability of these deep learning models and have recently emerged as a serious threat to face recognition systems. Adversarial examples are generated by adding perturbations to the input images, which are usually imperceptible to humans but can cause the model to make incorrect predictions. The majority of research on adversarial attacks~\cite{papernot2016limitations, szegedy2013intriguing, goodfellow2014explaining} presumes that the attacker can directly input the digitally generated adversarial example into the machine learning model. Such attacks are typically referred to as digital domain attacks. However, this assumption does not hold in the case of anti-spoofing, where the system is designed to work in the physical world. 

\begin{figure}[]
  \centering
   \includegraphics[width=1.0\linewidth]{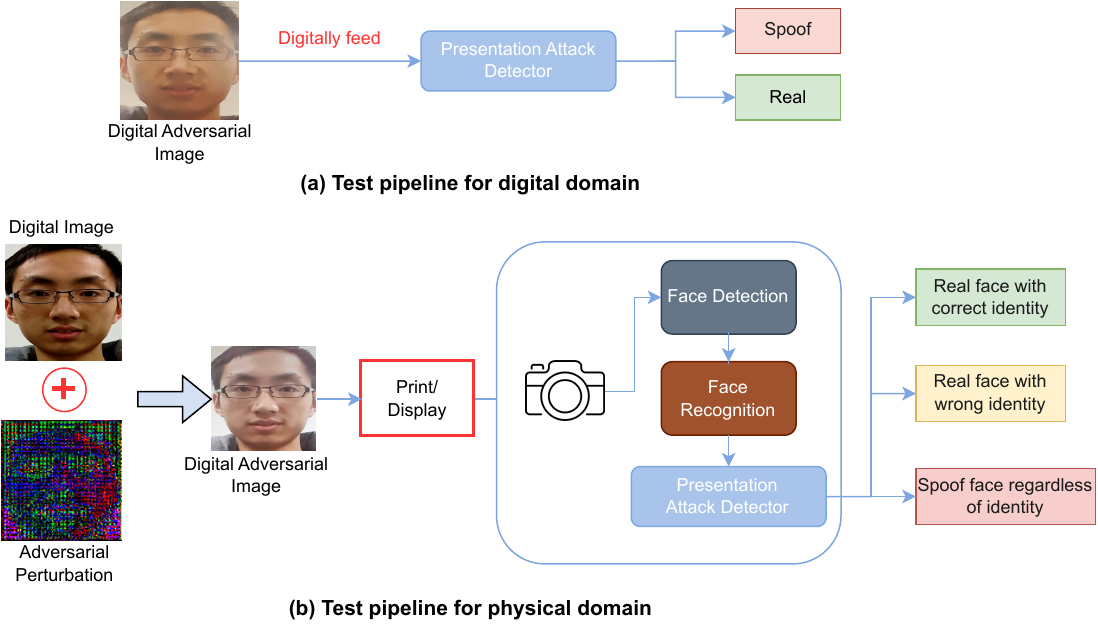}
   \caption{Experimental pipelines to evaluate the performance of the adversarial attacks. (a) shows the pipeline used when we attack a PAD in the digital domain, and (b) shows our testing pipeline in a physical domain. The digital image has to undergo two transformations and has to be effective after distortions are introduced in these processes.}
   \label{fig:flowchart}
\vspace*{-3mm}
\end{figure}

\begin{figure}[t]
  \centering
\includegraphics[width=1.0\linewidth]{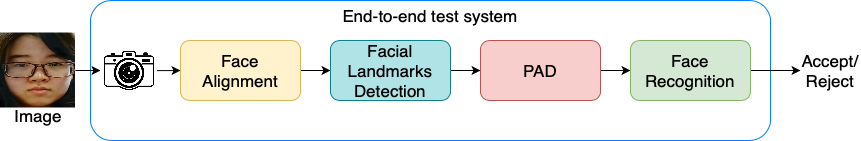}
   \caption{A typical Face authentication pipeline. Face PAD acts as a gatekeeper to face recognition module.}
   \label{fig:fig4}
\vspace*{-3mm}
\end{figure}

Adversarial attacks in the physical domain have gained significant attention in recent times due to their practicality and complexity. To attack the face anti-spoofing system in a physical world setting, the spoof image created by the attacker must be printed or displayed in the real world and then captured by the system's camera. This process of converting digital images to physical and then back to digital is called image rebroadcast~\cite{agarwal2018diverse}. The changes made to the image during this rebroadcast process help the anti-spoofing detector to recognize that the digital image is fake by looking exactly for the spoofing artifacts introduced during the rebroadcast process and prevent unauthorized access to the system. As a new spoofing pattern may be introduced after the attack, adversarial attacks need to act in a pre-emptive manner. Therefore, it is challenging to create an adversarial example that can effectively attack an anti-spoofing system in a physical domain setting. We show the difference between a physical and digital domain attack in Figure~\ref{fig:flowchart}.

After identifying the challenges associated with physical attacks, we present \modelname, an automated method to create adversarial face images. \modelname uses a Conditional Generative Adversarial Network to simulate 
presentation attacks and generate adversarial images that can fool state-of-the-art PADs in a physical domain attack setting. Our proposed method, \modelname, generates adversarial face images that mimic the process of physical presentation attacks, such as print and replay attacks. When a live image is passed through \modelname, it simulates the printing and displaying process to create an adversarial image that retains the characteristics of a printed or displayed image but is classified as real when passed through a spoof classifier. Moreover, \modelname ensures that the identity of the original face is preserved. The objective of \modelname is to incorporate the properties of physical adversarial attacks into digital adversarial attacks. The contributions of the
paper can be summarized as follows: 
\begin{enumerate}
    \item We design an identity preservation regularization term to enhance the identity preserving capability of a cycleGAN and name it IdGAN. IdGAN, given a real image, can generate a printed or replayed spoof version of it by preserving identity.
    
    \item We propose \modelname, a generative adversarial network trained to generate perturbations that are robust to distortions introduced to an image during physical transformations.

    \item A systematic mathematical formulation for the problem of generation of adversarial physical perturbation and modeling it as the learning objective of a deep generative model.

    \item We show that AdvGen is a more effective use of generating robust physical adversarial perturbations by  comparing it against  four datasets: SiW~\cite{zhang2019dataset}, MSU-MFSD~\cite{wen2015face}, Replay-Attack~\cite{chingovska2012effectiveness} and OULU-NPU~\cite{boulkenafet2017oulu}. (Figure \ref{fig:finalresult}).


\end{enumerate}

\section{Related Works}\label{related_works}
\noindent{\bf Adversarial Attacks}
 Many adversarial attack algorithms have indicated that deep learning models are broadly vulnerable to adversarial samples. For white-box attacks, where the attacker has complete knowledge of the target model, including its architecture and parameters, the gradient-based approaches \cite{goodfellow2014explaining, carlini2017towards, madry2017towards, dong2018boosting, dong2019efficient, brendel2017decision, wang2021enhancing} can be conducted by adding adversarial perturbations to the pixels of the original images, where all the perturbations are derived from the back-propagation gradients regarding the adversarial constraints. For black-box attacks, where the attacker has limited knowledge of the target model and must make queries to the model to infer its behavior in order to craft an effective attack, one interesting direction is to utilize a substitute/surrogate model to perform transfer-based attacks. Recent works \cite{zhong2020towards, xie2019improving, dong2019evading} claim that input diversity can further boost attack transferability. In the image classification domain, semi-whitebox approaches based on Generative Adversarial Networks (GANs) rely on softmax probabilities \cite{xiao2018generating, wang2019gan, song2018constructing, yang2021attacks}. Compared to digital attacks, physical attacks require much larger perturbation strengths to enhance the adversary’s resilience to various physical conditions such as lightness and object deformation~\cite{athalye2018synthesizing, xu2020adversarial}. Min-max optimization problem and transferability phenomenon are being explored for adversarial training~\cite{brendel2017decision,tramer2017space}. These explorations focus mostly on the region around natural examples where the loss is (close to) linear.

\noindent{\bf Generative Adversarial Networks (GANs)}
 Generative Adversarial Networks \cite{goodfellow2020generative} are now being used in a wide variety of applications. These include image synthesis applications \cite{radford2015unsupervised, denton2015deep}, style transfer \cite{ulyanov2016texture, johnson2016perceptual, gatys2016image}, image-to-image translation \cite{isola2017image, zhu2017unpaired}, and representation learning \cite{radford2015unsupervised, salimans2016improved, mathieu2016disentangling}. Previous studies with GAN have shown that it is possible to generate high-resolution images up to 1024 × 1024 resolution in various domains such as the human face, vehicles, and animals \cite{karras2017progressive, karras2019style}. In  \cite{goodfellow2014explaining} proposes a Fast Gradient Sign Method (FGSM) to generate adversarial examples. It computes the gradient of the loss function with respect to pixels and moves a single step based on the sign of the gradient. While this method is fast, using only a single direction based on the linear approximation of the loss function often leads to sub-optimal results.

 \noindent{\bf Adversarial Attacks on Face Recognition}
Current adversarial face synthesis methods include works by AdvFaces \cite{deb2020advfaces}, which learns to perturb the salient regions of the face, unlike FGSM \cite{goodfellow2014explaining} and PGD \cite{madry2017towards}, which perturbs every pixel in the image and  image is generated by gradient-based methods. LatentHSJA \cite{na2022unrestricted} manipulates the latent vectors for fooling the classification model, and \cite{zhang2020adversarial} which crafts replay-attack only to fool CNN-based face recognition system. Methods that rely on white-box manipulations of face recognition models are discussed first here. Bose et al. craft adversarial examples by solving constrained optimization such that a face detector cannot detect a face \cite{bose2018adversarial}. The adversarial eyeglasses can also be synthesized via generative networks \cite{sharif2019general}. But since these works are based on a white-box approach, it seems impractical in real-world scenarios. Dong et al. \cite{dong2019efficient} proposed an evolutionary optimization method for generating adversarial faces in black-box settings. This method requires at least $1,000$ queries to the target face recognition system before a realistic adversarial face can be synthesized. Song et al.~\cite{yang2021attacks} employed a conditional variation autoencoder GAN for crafting adversarial face images in a semi-whitebox setting. Here, they only focused on impersonation attacks and require at least five images of the target subject for training and inference.

\section{Methodology}\label{methodology}


\modelname consists of three components i) a simulator network that emulates printing and replaying input images, ii) a decomposition network that can decompose spoof faces into noise signal and live faces, and iii) a generator network supervised using a formulated loss to generate physical adversarial perturbations.

We formulate the problem of generating a robust physical adversarial perturbation as an optimization objective in Section 3.1. Then we describe the architecture of the simulator network in Section 3.2. In Section 3.3, we elaborate on modeling the formulated optimization objective using a Generative Neural network. 

\begin{figure}[t]
  \centering
   \includegraphics[width=0.9\linewidth,height=0.43\textheight]{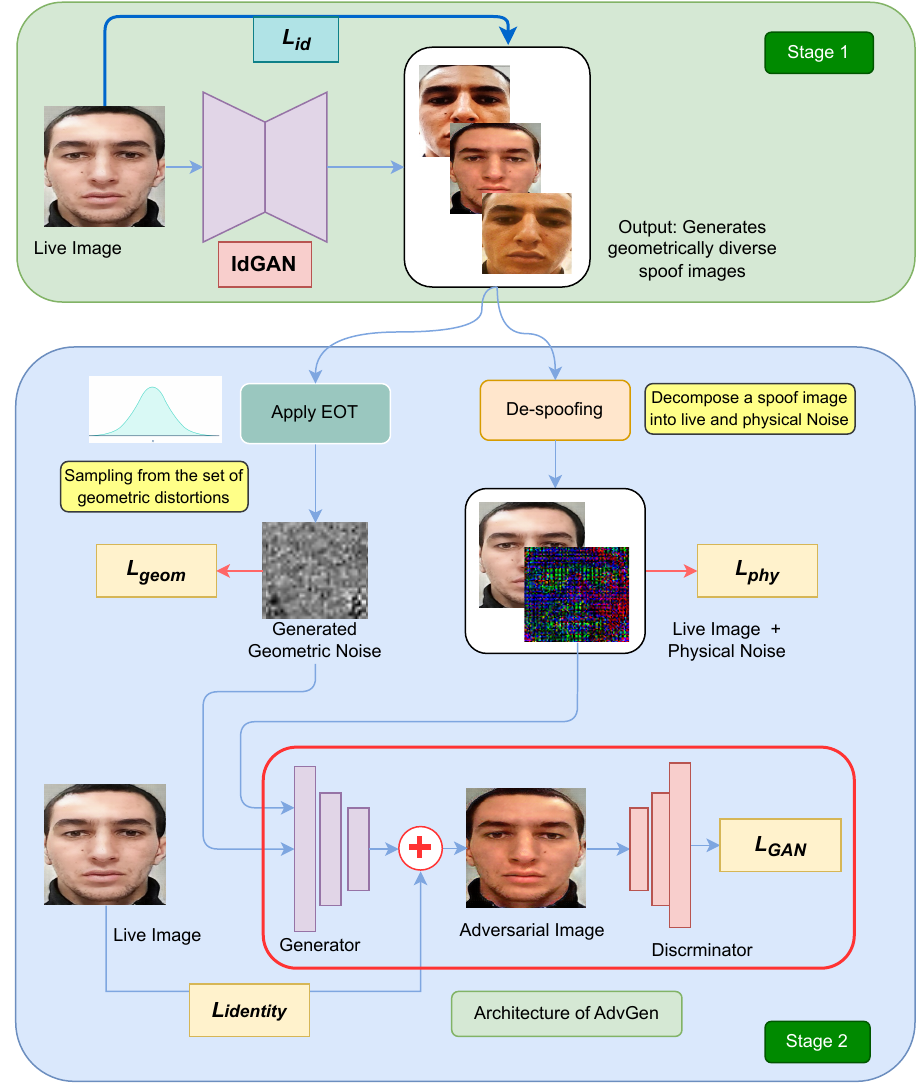}
   \caption{Synthesizing adversarial face images using \modelname consists of two stages: \textbf{Stage 1:} Training of \textit{IdGAN} which, given a live image, learns to generate geometrically diverse spoof images. These generated images produced by \textit{IdGAN} simulate printing and replay. \textit{Identity loss} is introduced as an identity regularizer to preserve the subject's identity in the generated images. \textbf{Stage 2:} We apply de-spoofing and EOT on the generated spoof images to get the physical and geometric noises. These are fed into AdvGen's generator to generate the adversarial perturbation. The generated image from \modelname is robust to physical as well as geometric distortions.}
   \label{fig:architecture}
\vspace*{-3mm}
\end{figure}

\subsection{Problem Formulation}

First, we formulate the creation of an adversarial image in the digital domain, and then we modify it to the physical domain.

Let $\mathcal{I}$ denote an input image and $l_{true}$ its corresponding label. Let $l_{target} \neq  l_{true}$ be the target label of the attack. Let $f\left( \cdot  \right)$ denote the output of the target neural network. The process of generating an adversarial perturbation $\delta$ involves solving the following optimization problem:
\begin{equation}
\begin{aligned}
   \arg \min_{\delta} \mathcal{L}(f(\mathcal{I} + \delta), l_{target}), \\
   \textrm{subject to} \parallel\delta\parallel_{p} < \epsilon \label{eq:equation1}
\end{aligned}
\end{equation}
where $\mathcal{L(\cdot)}$ is the neural network's loss function, and ${\parallel\cdot\parallel_{p}}$ denotes the $L_{p}$-norm.  To solve the above-constrained optimization problem efficiently, we reformulate it in the Lagrangian-relaxed form:
\begin{equation}
\begin{aligned}
   \arg \min_{\delta} \mathcal{L}(f(\mathcal{I} + \delta), l_{target}) + \lambda\parallel\delta\parallel_{p} \label{eq:equation2}
\end{aligned}
\end{equation}
where $\lambda$ is a hyper-parameter that controls the regularization of the distortion $\parallel\delta\parallel_{p}$.


In a physical domain setting, we denote a spoof image as $\mathcal{I}_{s}$. The spoof detection network is not fed directly with $\mathcal{I}_{adv} = \mathcal{I}_{s} + \delta^{\ast}$ ($\delta^{\ast}$ is the optimal digital perturbation obtained by using Eq. \ref{eq:equation2} with its physically recaptured version 
$\mathcal{I}_{r} = \mathcal{P}(\mathcal{I}_{adv}) = \mathcal{P}(\mathcal{I}_{s} + \delta^{\ast})$ 
where we use $\mathcal{P}(\cdot)$ to denote the physical broadcasting and recapture procedure. $\mathcal{P}(\cdot)$ is capable of destroying the effect of $\rho^{\ast}$.

In order to ensure that the perturbation remains effective even after the image has been rebroadcasted, it is important to consider the possible transformations that the image may undergo during this process. This will allow us to create a robust perturbation that can withstand these transformations. $\mathcal{T}$ denotes the set of all transformations in the physical process.
Perturbation $\rho$ can be obtained by optimizing the average loss over $\mathcal{T}$, 
\begin{equation}
\begin{aligned}
    \arg\min_{\rho} \mathbb{E}_{t\sim\mathcal{T}} [\mathcal{J}(f_s(t(\mathcal{I}) + \rho), l_{target})] + \lambda\parallel\rho\parallel_{p} \label{eq:equation3}
\end{aligned}
\end{equation}
Here $f_s$ denotes the output of a face presentation attack detector for a transformed image $\mathcal{I}$ after applying a broadcasting transform $t$ selected from a set of physical transforms $\mathcal{T}$ and then applying a perturbation $\rho$ obtained using Eq.~\ref{eq:equation3}.

\subsection{Physical Simulator Network}

We train \textit{\textbf{IdGAN}}, an architecture derived from CycleGAN, to learn the simulation from real to spoof. This network learns to add physical and geometrical perturbations to an input image. It has two benefits: i) the simulated image will be useful in the next stage of attack generation, ii) This network is trained on data exposed to physical augmentations(rotation, random crop, resize, etc.), making the network capable of generating spoof images with physical variations. \\
\textbf{Generators:} The network consists of two generators $\mathcal{G}_{rs}$ and $\mathcal{G}_{sr}$. 
Generators are based on Convolution based encoder-decoder architectures and generate a feature representation of the input image $\mathcal{I}_r$, and the decoder generates the corresponding presentation attack variants of the input $\mathcal{I}_r$. 
The discriminators $\mathcal{D}_r$ and $\mathcal{D}_s$ distinguish between the captured examples and the generated samples by the generators.
The network is trained using three types of losses:
\begin{enumerate}
\item \textbf{Identity Regularizer:} The generated image should preserve the identity of the input. This would be a critical component in the adversarial attack generation. We introduce an identity-preserving regularization term to CycleGAN. The network, at every iteration, tries to preserve identity by minimizing the cosine similarity between the face embeddings of the generated image and the input image. The face embeddings are generated using a pretrained ArcFace~\cite{deng2019arcface}. The identity regularizer is defined as,
    \begin{equation}
    \begin{split}
        \mathcal{L}_{id}(\mathcal{G}_{rs}, \mathcal{G}_{sr}, \mathcal{I}_r, \mathcal{I}_s) & = \mathbb{E}_{x}[1-\mathcal{F} [\mathcal{G}_{sr}(\mathcal{G}_{rs}(\mathcal{I}_r)), \mathcal{I}_r]] \\ & + \mathbb{E}_{x}[1-\mathcal{F}[\mathcal{G}_{rs}(\mathcal{G}_{sr}(\mathcal{I}_s)),\mathcal{I}_s]] 
    \label{eq:equation4}
    \end{split}
    \end{equation}  
\vspace{-5mm}   

\begin{figure}[t]
  \centering
   \includegraphics[width=0.8\linewidth]{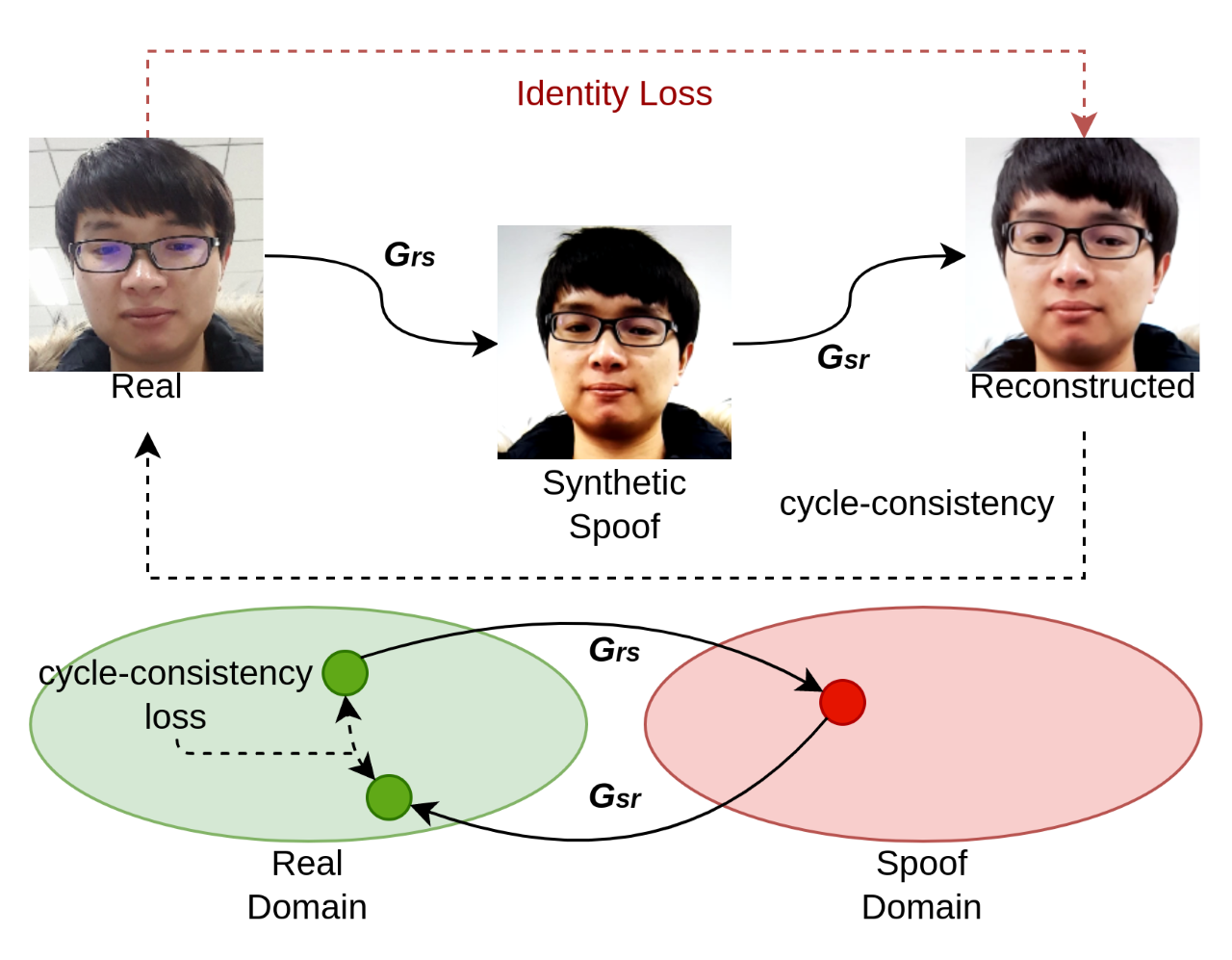}
   \caption{Loss terms used to train \textbf{IdGAN}. along with conventional $\mathcal{L}_{adv}$ and $\mathcal{L}_{cycle}$, we introduce a $\mathcal{L}_{id}$ to preserve identity in the generated image, which is a crucial step for the stage 2.}
   \label{fig:cycleloss}
\vspace*{-3mm}
\end{figure}
\item \textbf{Adversarial Loss}: Adversarial loss creates a 2-player adversary between the generator and discriminator, leading to better training through competition. An MSE-based adversarial loss is used and defined as,
\begin{equation}
\begin{split}
        \mathcal{L}_{adv}(\mathcal{G}_{rs}, \mathcal{D}_s,& \mathcal{I}_r, \mathcal{D}_r) =  \mathbb{E}_{\mathcal{I}_s\sim p_{data}(\mathcal{I}_s)}log[\mathcal{D}_{s}(\mathcal{I}_s)] + \\
        & \mathbb{E}_{\mathcal{I}_r\sim p_{data}(\mathcal{I}_r)}log[1 -\mathcal{D}_{s}(\mathcal{G}_{rs}(\mathcal{I}_r))] \\
        \mathcal{L}_{adv}(\mathcal{G}_{sr}, \mathcal{D}_r,& \mathcal{I}_s, \mathcal{D}_s) =  \mathbb{E}_{\mathcal{I}_r\sim p_{data}(\mathcal{I}_r)}log[\mathcal{D}_{r}(\mathcal{I}_r)] + \\
        & \mathbb{E}_{\mathcal{I}_s\sim p_{data}(\mathcal{I}_s)}log[1 -\mathcal{D}_{r}(\mathcal{G}_{sr}(\mathcal{I}_s))] \\
        \mathcal{L}_{adv} =  \mathcal{L}_{adv}(&\mathcal{G}_{rs}, \mathcal{D}_s, \mathcal{I}_r, \mathcal{D}_r) + \\
        & \mathcal{L}_{adv}(\mathcal{G}_{sr}, \mathcal{D}_r, \mathcal{I}_s, \mathcal{D}_s) 
\end{split}
\label{eq:equation5}
\end{equation}
\vspace*{-3mm}
    \item \textbf{Cycle Consistency Loss}: Adversarial loss leaves the learning unconstrained. Hence the Cycle Consistency Loss is added as a regularization term to the generator's objectives shown in Figure \ref{fig:cycleloss}. This loss is defined as, 
    \begin{equation}
    \begin{split}
        & \mathcal{L}_{cyc}(\mathcal{G}_{rs},  \mathcal{I}_r) = \mathbb{E}_{\mathcal{I}_r\sim p_{data}(\mathcal{I}_r)}[\left\| \mathcal{G}_{sr}(\mathcal{G}_{rs}(\mathcal{I}_r))-\mathcal{I}_r \right\|_1] \\
        & \mathcal{L}_{cyc}(\mathcal{G}_{sr},  \mathcal{I}_s) = \mathbb{E}_{\mathcal{I}_s\sim p_{data}(\mathcal{I}_s)}[\left\| \mathcal{G}_{rs}(\mathcal{G}_{sr}(\mathcal{I}_s))-\mathcal{I}_s \right\|_1] \\
        & \mathcal{L}_{cycle} = \mathcal{L}_{cyc}(\mathcal{G}_{rs},  \mathcal{I}_r) + \mathcal{L}_{cyc}(\mathcal{G}_{sr},  \mathcal{I}_s)
    \end{split}
    \label{eq:equation6}
    \end{equation}

Here $\left\| \cdot \right\|_1$ denotes $\mathcal{L}_1$ norm
\end{enumerate}
Finally, IdGAN is trained using the following objective,
\begin{equation}
    \begin{split}
        \mathcal{L} = \mathcal{L}_{adv} + \lambda_{cycle} \times \mathcal{L}_{cycle} + \lambda_{id} \times \mathcal{L}_{id}
    \label{eq:equation7}
    \end{split}
    \end{equation}
\subsection{Modelling the Physical Transformation}
A real image $\mathcal{I}$ undergoes physical transformations such as color distortion and display, printing, and imaging artifacts to become a spoof image~\cite{jourabloo2018face}. In addition, the presenter may wish to introduce geometric distortions like rotation, capture distance, folding the presentation medium, etc. These distortions need to be carefully modeled. 
To generate the perturbation, we use a generative neural network to model the optimization problem. \modelname is optimized over the formulated loss. Figure \ref{fig:architecture} outlines the proposed architecture. \modelname consists of a generator $\mathcal{G}$, a discriminator $\mathcal{D}$, a spoof noise synthesiser $\mathcal{S}$ and a geometric distortion sampler $\mathcal{F}$. Together these modules model every necessary component in the formulated objective. \\
\\
\textbf{Generator} The generator $\mathcal{G}$ of \modelname takes in an input image $ x \in \mathcal{X}$ and generates a perturbation $\mathcal{G}(x)$. In order to maintain the original visual quality of the input image and avoid generating a completely new face image, the generator produces an additive perturbation that is applied to the input image as $x+\mathcal{G}(x)$. The generator's loss has the following components:\\
\\
\textbf{Physical Perturbation Hinge Loss:} To generate perturbations that include physical distortions, we use a pretrained noise decomposition network~\cite{jourabloo2018face}. It is in the synthesized spoof image from \modelname, and returns decomposed physical noise and live faces. This synthesized noise serves as the perturbation to be added to the real image. This noise is an unbounded physical noise. Hence we introduce this noise to the generation pipeline using a soft hinge loss on the $\mathcal{L}_2$ norm bounding the amount of physical noise introduced by ~\cite{carlini2017towards,liu2016delving} formulated as:
\begin{equation}
    \begin{split}
        \mathcal{L}_{phy} = \mathbb{E}_{x}[\max( \epsilon_1, \left\|\mathcal{P}hy(x)\right\|_2)]
    \end{split}
    \label{eq:equation8}
    \end{equation}
$\epsilon_1$ is a user-specific bound on the added perturbation and $\mathcal{P}hy(\cdot)$ denotes physical noise from the decomposition network.\\

\textbf{Geometric Distortion Hinge Loss:} Presentation of a physical medium is always subject to geometric distortions such as rotation, zooming, folding, etc., due to human errors. To make the attack robust to geometric distortions, \modelname is trained with geometric augmentations to generate spoof images with diverse geometric variations. To model these distortions, Expectation over Transforms(EOT)~\cite{athalye2018synthesizing} is applied over the generated spoof images. Modeling these transformations diversifies the set of physical transforms modeled by the generator. The generated geometric perturbation is controlled using a geometric hinge loss 
\begin{equation}
    \begin{split}
        \mathcal{L}_{geom} = \mathbb{E}_{x}[\max( \epsilon_2, \left\|\mathcal{G}eom\right\|_2)]
    \end{split}
    \label{eq:equation9}
\end{equation}
$\epsilon_2$ is a user-specific bound on the added perturbation and $\mathcal{G}eom(\cdot)$ denotes geometric perturbation obtained from EOT.\\

\textbf{Identity Regularizer Loss:} The perturbation must preserve the identity of the target. We introduce an identity regularizer to the generator loss, which maximizes the cosine similarity between the identity embeddings obtained from a pretrained ArcFace~\cite{deng2019arcface} matcher. We define it as,
\begin{equation}
        \mathcal{L}_{identity} = \mathbb{E}_{x}[1-\mathcal{F}(x, x + \mathcal{G}(x))]
    \label{eq:equation10}
\end{equation} 
\textbf{Discriminator}: We introduce a discriminator $\mathcal{D}$ which distinguishes between the generated samples $x + \mathcal{G}(x)$ and the corresponding real sample $x$. This Discriminator is based on PatchGAN and projects the input to a patch-based matrix where each value in the matrix corresponds to the score of the particular patch's discriminative score. trained using the adversarial loss: 
\begin{equation}
        \mathcal{L}_{GAN} = \mathbb{E}_{x}[\log\mathcal{D}(x)] + \mathbb{E}_{x}[\log(1-\mathcal{D}(x + \mathcal{G}(x))]
    \label{eq:equation11}
\end{equation} 

\modelname is trained to generate identity-preserving physical perturbation in an end-to-end on the following objective:
\begin{equation}
    \begin{split}
        \mathcal{L} = 
        \lambda_{phy} &\times \mathcal{L}_{phy} +
        \lambda_{geom} \times \mathcal{L}_{geom} + \\
        & \lambda_{identity} \times \mathcal{L}_{identity} + 
        \lambda_{GAN} \times \mathcal{L}_{GAN}
    \label{eq:equation12}
    \end{split}
    \end{equation}

\section{Experiments}\label{sec:experiments}
In this section, we first introduce the datasets used and the experimental setup. Then we evaluate the performance of our framework in different settings and explain the evaluation metrics: 

\subsection{Datasets and Baselines}
\begin{table*}[t]
\renewcommand*{\arraystretch}{1.3}
\centering
\begin{tabular}{l@{\hspace{1.5cm}}l@{\hspace{1.5cm}}l@{\hspace{1.5cm}}l@{\hspace{1.5cm}}l@{\hspace{1.5cm}}l@{\hspace{1.5cm}}}
\hline
\multicolumn{6}{c}{\textit{\textbf{Attack Success Rate on OULU-NPU(\%) and SSIM after attack}}}                     \\ \hline
\textbf{}    & BIM~\cite{kurakin2018adversarial}         & EOT~\cite{athalye2018synthesizing}      & $RP_2$~\cite{eykholt2018robust}      & D2P~\cite{jan2019connecting}      & \textbf{Ours}       \\ \hline
CDCN~\cite{yu2020searching}         & 41.19       & 55.82    & 63.12       & 68.37    & \textbf{81.02}      \\
CDCNpp~\cite{zhang2021celeba}      & 37.47       & 51.61    & 59.39       & 64.26    & \textbf{78.22}      \\
C-CDN~\cite{yu2021dual}       & 38.38       & 51.58    & 60.83       & 65.49    & \textbf{79.34}      \\
DC-CDN~\cite{yu2021dual}     & 39.95       & 53.83    & 61.36       & 66.03    & \textbf{80.55}      \\ \hline
SSAN-M~\cite{wang2022domain}       & 40.06       & 52.02    & 61.40       & 65.27    & \textbf{80.42}      \\
SSAN-R~\cite{wang2022domain}       & 34.54       & 49.83    & 57.03       & 61.79    & \textbf{75.15}      \\ \hline
DBMNet~\cite{jia2021dual}       & 38.78       & 52.69    & 59.89       & 62.74    & \textbf{79.63}      \\
STDN~\cite{liu2020disentangling}         & 40.92       & 53.93    & 61.67       & 63.29    & \textbf{80.98}      \\
Meta-FAS~\cite{cai2022learning}     & 35.38       & 47.67    & 57.25       & 59.53    & \textbf{76.19}      \\ 
De-Spoofing~\cite{jourabloo2018face}  & 46.44       & 58.43    & 65.41       & 68.66    & \textbf{84.67}      \\ \hline
\textit{\textbf{SSIM in [0,1]}}         & \textit{\textbf{0.64}}       & \textit{\textbf{0.38}}    & \textit{\textbf{00.32}}       & \textit{\textbf{0.45}}    & \textit{\textbf{0.98}}      \\ \hline
\end{tabular}
\vspace*{1.5mm}
\caption{Comparison of attack success rates on different models and ours using four different datasets.}
\label{tab:compare}
\end{table*}
We train \modelname on OULU-NPU~\cite{boulkenafet2017oulu} and test on SiW~\cite{zhang2019dataset}, MSU-MFSD~\cite{wen2015face}, Replay-Attack~\cite{chingovska2012effectiveness} and OULU-NPU~\cite{boulkenafet2017oulu}\footnote{We train on training and validations sets of Protocol 1 of OULU-NPU and test on the corresponding test set} datasets.
 \textbf{OULU-NPU~\cite{boulkenafet2017oulu}} face presentation attack detection database contains 4,950 real access and attack videos belonging to 55 different subjects.
\textbf{SiW~\cite{zhang2019dataset}} contains 4,478 15s long videos for 165 subjects. For each subject, there are eight live and up to 20 spoof videos.
\textbf{MSU-MFSD~\cite{wen2015face}} contains 280 video recordings of genuine and attack faces for 35 individuals.
\textbf{Replay-Attack~\cite{chingovska2012effectiveness}} consists of 1300 video clips of photo and video attacks for 50 clients under different lighting conditions.

We compare our proposed method with four \sota physical attack generation methods BIM~\cite{kurakin2018adversarial}, EOT~\cite{athalye2018synthesizing}, $RP_2$~\cite{eykholt2018robust}, D2P~\cite{jan2019connecting}. To compare our method's effectiveness in the physical vs. digital domain, we implement four standard digital adversarial attacks FGSM~\cite{goodfellow2014explaining}, PGD~\cite{madry2017towards}, BIM~\cite{kurakin2018adversarial}, and Carlini \& Wagner~\cite{carlini2017towards}. We use TorchAttack's~\cite{kim2020torchattacks} implementations of the above methods by perturbing the necessary parameters to generate effective attacks. To establish the effectiveness and generalizability of our proposed attack across different spoof detection models, we compare the ASR of our generated images from OULU-NPU across ten state-of-the-art face anti-spoofing models in Table \ref{tab:compare}. 

\subsection{Evaluation Metrics}

By comparing our network against state-of-the-art baselines, we quantify the adversarial attacks' effectiveness via i) attack success rate (ASR) and ii) structural similarity (SSIM) \cite{wang2004image}.

The attack success rate (ASR) is computed as
\begin{equation}
    ASR = \frac{\text{No. of attacks classified as real}}{\text{Total number of attacks}} \times 100 \%
    \label{eq:equation13}
\end{equation}
To quantify the effectiveness of the generated adversarial images with the input image, we compute the Structural Similarity Index (SSIM) metric calculated between the adversarial image and the real image as proposed in research\cite{wang2004image}:

\subsection{Experimental Setup}

All experiments are conducted on print and replay attack scenarios. We use an HP Smart Tank 580 printer to print all the images. For display, we use two mediums, MacBook Pro (Intel Iris Plus Graphics 640 1536 MB) and Redmi K20 pro (Super AMOLED, HDR10 display). All images are captured from a distance ranging from 20cm to 40cm.

To validate the effectiveness of our developed attack method, we deploy four state-of-the-art face anti-spoofing methods to a streamlit app. The app takes a real-time feed and returns the predicted identity of the person along with spoof/live prediction along with its confidence. 

We create a test set of 300 images per dataset comprising different identities. From OULU-NPU, we sample 20 identities; from SiW, we sample 50 identities; from REPLAY-ATTACK, we sample 15 identities; from MSU-MFSD, we sample 15 identities. The sampled images are manually handpicked to ensure that maximum diversity is covered in terms of variations. To validate results for EOT, we manually perform physical distortions like rotation on the print and replay displays, change of brightness in the replay attacks, and folding the presentation medium in print attacks.

\subsection{Experimental Settings}
We use ADAM optimizers with ${\beta_1 = 0.5}$ and ${\beta_2 = 0.9}$. Each mini-batch consists of 1 face image. We train \modelname for 100 epochs with a fixed learning rate of 0.0002. We also use identity loss with parameters $\lambda_{i}$ = 1.0. We train two separate models for print and video-replay attacks. A unified model for both attacks is also trained with the same hyperparameters. We iteratively perform FGSM over \modelname with $\epsilon$ = 0.1. All experiments are conducted using PyTorch.

\section{Results and Analysis}\label{sec:results}
\subsection{Effectiveness in Physical Domain}
\begin{table}[ht]
\renewcommand*{\arraystretch}{1.2}
\centering
\begin{tabular}{ccc} \hline
\multicolumn{3}{c}{\textbf{Attack Success Rate (\%)}} \\ \hline
\textbf{}    & Digital Domain    & Physical Domain    \\ \hline
BIM~\cite{kurakin2018adversarial}           & 98.04                                                         & 41.22                                                            \\
FGSM~\cite{goodfellow2014explaining}          & 75.32                                                        & 23.13                                                         \\
GA            & 79.56                                                        & 26.92                                                         \\
IGSA          & 100.00                                                          & 34.22                                                          \\
IGA           & 99.64                                                         & 31.48                                                          \\
PGD~\cite{madry2017towards}           & 98.63                                                         & 36.42                                                          \\
\textbf{AdvGen} & {\color[HTML]{333333} \textbf{100}}                           & {\color[HTML]{333333} \textbf{81.02}}                          \\ \hline
\end{tabular}
\vspace*{1.5mm}
\caption{Performance of \sota adversarial attack methods in the digital and physical domain. }
\label{tab:phy-eff}
\end{table}

\begin{figure}[]
  \centering
   \includegraphics[width=1.0\linewidth]{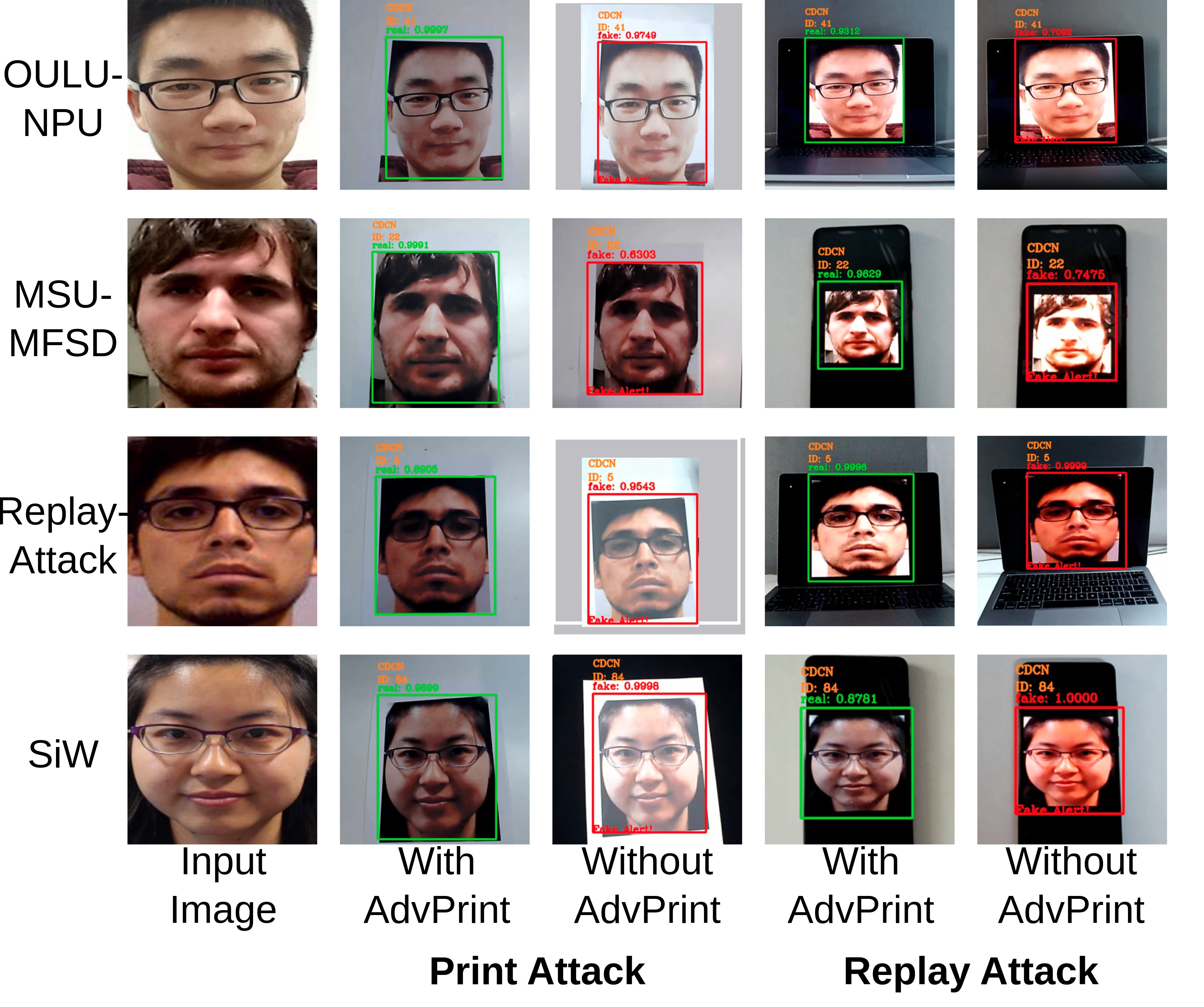}
   \caption{Experimental pipelines to evaluate the performance of the attacks. (a) shows the pipeline used when we attack a PAD in the digital domain, and (b) shows our testing pipeline in a physical world setting.}
   \label{fig:comparison}
\vspace*{-3mm}
\end{figure}

To evaluate the effectiveness of the proposed method in the physical domain, we perform a digital attack using conventional attack strategies and our method on the test set of 300 images curated from OULU-NPU. Then the adversarial images are printed and presented physically to a presentation attack detector. 
The performance of all attacks is optimal in the digital domain but significantly drops when transferred to the physical domain, as demonstrated in Table~\ref{tab:phy-eff}. The ASR of the standard methods is less than 50 in the physical domain, while our method clearly outperforms these values. These empirical results clearly demonstrate that including physical spoofing noise makes the attack robust to transformations incurred through physical processes.

\subsection{Comparison Studies}
\begin{figure*}[h]
  \centering
   \includegraphics[width=0.9\linewidth]{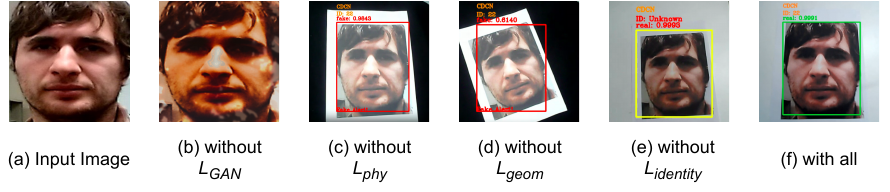}
   \caption{Variants of \modelname trained without GAN loss, physical perturbation hinge loss, geometric distortion hinge loss, and identity loss, respectively.}
   \label{fig:ablation}
\vspace*{-3mm}
\end{figure*}

In Table \ref{tab:compare}, we present the findings from our comparative studies against state-of-the-art physical adversarial attack methods. Compared to the state-of-the-art methods, our method is significantly better at generating robust attacks in terms of achieved ASR. In terms of structural similarity, our method stands out in preserving visual information in the generated image and outperforms the other methods. Our method learns to generate imperceptible noise signals at locations on the face that are not significant for identity recognition. BIM~\cite{kurakin2018adversarial} iteratively generates perturbations on the input image, hence preserving visual features to some extent, but the ASR on the generated images is low because of its inability to model physical perturbations. Attack images generated using EOT, $RP_2$, and D2P have higher ASR by virtue of their design to address generic physical distortions in their noise modeling. They are able to generate physically robust attacks as compared to BIM, but these are not specifically physical perturbations introduced on a face image due to physical transformations like printing or display on a screen. Our method models this noise and hence is better at modeling. 


\subsection{Effectiveness with Geometric Distortions}
\begin{figure}[h]
  \centering
   \includegraphics[width=1.0\linewidth]{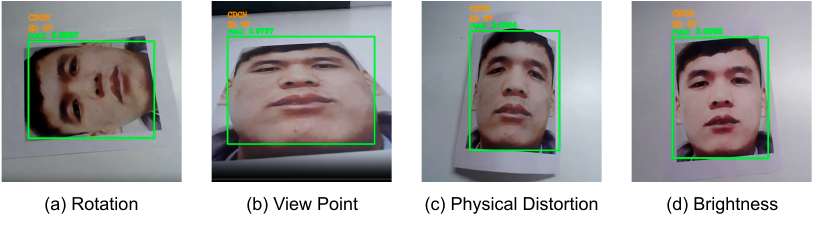}
   \caption{Effectiveness of \modelname after applying geometric distortions. Adversarial image is classified as real (a) after rotation, (b) changing the viewpoint of the camera, (c) applying physical distortions, like folding the image, and (d) changing the brightness level of the setup.  }
   \label{fig:eot}
\vspace*{-3mm}
\end{figure}

In physical presentations, geometric distortions like capturing viewpoint, rotation, scaling, and perspective changes of the display medium and folding of the printed medium are unavoidable. Being trained on distortions sampled by Expectation Over Transformation(EOT)~\cite{athalye2018synthesizing}, our method is robust to geometric distortions like viewpoint changes, rotation, and brightness. Figure \ref{fig:eot} demonstrates the effectiveness of our methods through various geometric distortions.
\subsection{Ablation Study}
\begin{figure}[h]
  \centering
   \includegraphics[width=0.8\linewidth]{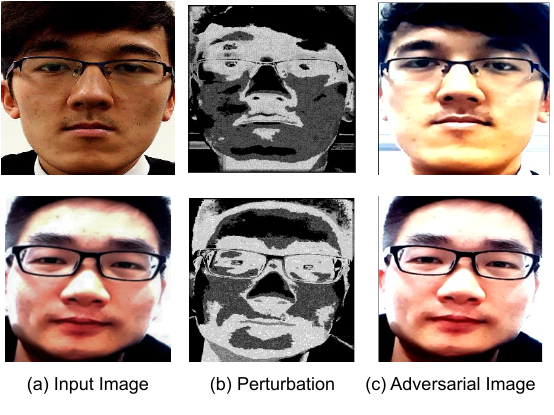}
   \caption{Visualization of the generated perturbation. (a) shows the input image, which can be live or spoof, (b) the locations of the input face resulting in perturbation we get from \modelname, and (c) shows the final adversarial image.}
   \label{fig:perturbation}
\vspace*{-3mm}
\end{figure}

AdvGen is trained using four loss terms, each contributing to one component to be added to the generated perturbation. To analyze the importance of each module, we train four variants of AdvGen for comparison by dropping $\mathcal{L}_{phy}$, $\mathcal{L}_{geom}$, $\mathcal{L}_{identity}$ and $\mathcal{L}_{GAN}$ and show results in Figure \ref{fig:ablation}. Without a discriminator, i.e., with $\mathcal{L}_{GAN}$, the visual quality of generated images is affected, and undesirable artifacts are introduced. Without a physical perturbation hinge $\mathcal{L}_{phy}$, the generated perturbation is not robust enough to physical transformation and gets classified as a "spoof."  Perturbations generated without being regulated by any geometric distortion $\mathcal{L}_{geom}$  fail even when even a small geometric distortion is performed. Without an identity regularizer, though, the generated perturbation is robust for a presentation attack generator but fails to pass the identity check. The generated perturbation by such a generator perturbs the identity. We conclude that to generate a perceptually realistic and robust perturbation, every component is necessary.


\section{Future Works}\label{future_works}
Focusing on the print and reply attack scenario, we proposed \modelname, which generates adversarial images to fool a face PAD. Below, we list a few points that we would like to pursue in the future: 

\begin{enumerate}
    \item Extending our attack to a scenario in which the attack is carried out by showing a 3D and paper mask, make-up, mannequin, etc., of the adversarial example to the authentication system.
    
    \item From the defender’s side, future research has to be performed to recover robustness against anti-spoofing and design new CNN-based face authentication systems capable of working in the presence of adversarial spoofing attacks.
    
    \item Having demonstrated the threats posed by replay and print attacks exploiting adversarial examples, we plan to propose a defense for such attacks. We will create a system that would be capable of working in the presence of such adversarial print and replay images.
\end{enumerate}

\section{Conclusion}\label{conclusion}
In this paper, we have created a physical attack on a CNN-based face authentication system that has an anti-spoofing module. We demonstrate that attacking an anti-spoofing face authentication system in the physical domain is more challenging and comes with additional difficulties than attacking systems in other application scenarios. Our new framework, called \modelname, can produce adversarial images that mimic a printing and replay procedure. Through experimentation, we have demonstrated that \modelname can generate synthetic adversarial prints that are capable of bypassing the Presentation Attack Detectors (PADs) and fooling a face recognition system, all while maintaining the subject's identity.

{\small
\bibliographystyle{ieee}
\bibliography{bibliography}
}

\end{document}